\documentclass[letterpaper]{article} 
\usepackage{aaai25}  
\usepackage{times}  
\usepackage{helvet}  
\usepackage{courier}  
\usepackage{xcolor}
\usepackage[hyphens]{url}  
\usepackage{graphicx} 
\usepackage{natbib}  
\usepackage{caption} 
\usepackage{algorithm}
\usepackage{algorithmic}
\usepackage{amsmath,amsfonts}
\usepackage{newfloat}
\usepackage{listings}
\usepackage{xcolor}

\DeclareCaptionStyle{ruled}{labelfont=normalfont,labelsep=colon,strut=off} 
\floatstyle{ruled}
\newfloat{listing}{tb}{lst}{}
\floatname{listing}{Listing}
\title{
Unified Coding for Both Human Perception and Generalized Machine Analytics with CLIP Supervision} 
\author{
Kangsheng Yin\textsuperscript{\rm 1}\textsuperscript{\rm 3}\thanks{These authors contributed equally.}, Quan Liu\textsuperscript{\rm 1}\textsuperscript{\rm 3}\footnotemark[1], Xuelin Shen\textsuperscript{\rm 1}\footnotemark[2] , Yulin He\textsuperscript{\rm 1}, Wenhan Yang\textsuperscript{\rm 2}\thanks{Corresponding author.} , Shiqi Wang\textsuperscript{\rm 4}
}

\affiliations{
\textsuperscript{\rm 1}Guangdong Laboratory of Artificial Intelligence and Digital Economy (SZ)\\
\textsuperscript{\rm 2}Peng Cheng Laboratory\\
\textsuperscript{\rm 3}Shenzhen University\\
\textsuperscript{\rm 4}City University of Hong Kong
}

\usepackage{bibentry}
\begin{document}

\maketitle

\begin{abstract}
The image compression model has long struggled with adaptability and generalization, as the decoded bitstream typically serves only human or machine needs and fails to preserve information for unseen visual tasks.
Therefore, this paper innovatively introduces supervision obtained from multimodal pre-training models and incorporates adaptive multi-objective optimization tailored to support both human visual perception and machine vision simultaneously with a single bitstream, denoted as Unified and Generalized Image Coding for Machine (UG-ICM). 
Specifically, to get rid of the reliance between compression models with downstream task supervision, we introduce Contrastive Language-Image Pre-training (CLIP) models into the training constraint for improved generalization.
Global-to-instance-wise CLIP supervision is applied to help obtain hierarchical semantics that make models more generalizable for the tasks relying on the information of different granularity.
Furthermore, for supporting both human and machine visions with only a unifying bitstream, we incorporate a conditional decoding strategy that takes as conditions human or machine preferences, enabling the bitstream to be decoded into different versions for corresponding preferences.
As such, our proposed UG-ICM is fully trained in a self-supervised manner, \textit{i.e.}, without awareness of any specific downstream models and tasks.
The extensive experiments have shown that the proposed UG-ICM is capable of achieving remarkable improvements in various unseen machine analytics tasks, while simultaneously providing perceptually satisfying images.
\end{abstract}

\begin{links}
    \link{Code}{https://github.com/YinKangSheng/UG-ICM}
\end{links}
\section{Introduction}
Recent years have witnessed the rise of machine-vision-oriented application scenarios, \textit{e.g.} autonomous driving, smart cities, and the Internet of Things.
Existing technologies need to shift towards meeting the demands of machine vision.
Therefore, on the coding side, the demand for machine-oriented coding also emerges.
Some preliminary attempts, \textit{e.g.}, feature coding~\cite{chenzhuo, chen2019lossy}, are implemented to compress the intermediate deep features, 
which satisfies only the machine vision needs by totally discarding low-level information.

\begin{figure}[t]
      \centerline{\includegraphics[width=8cm]{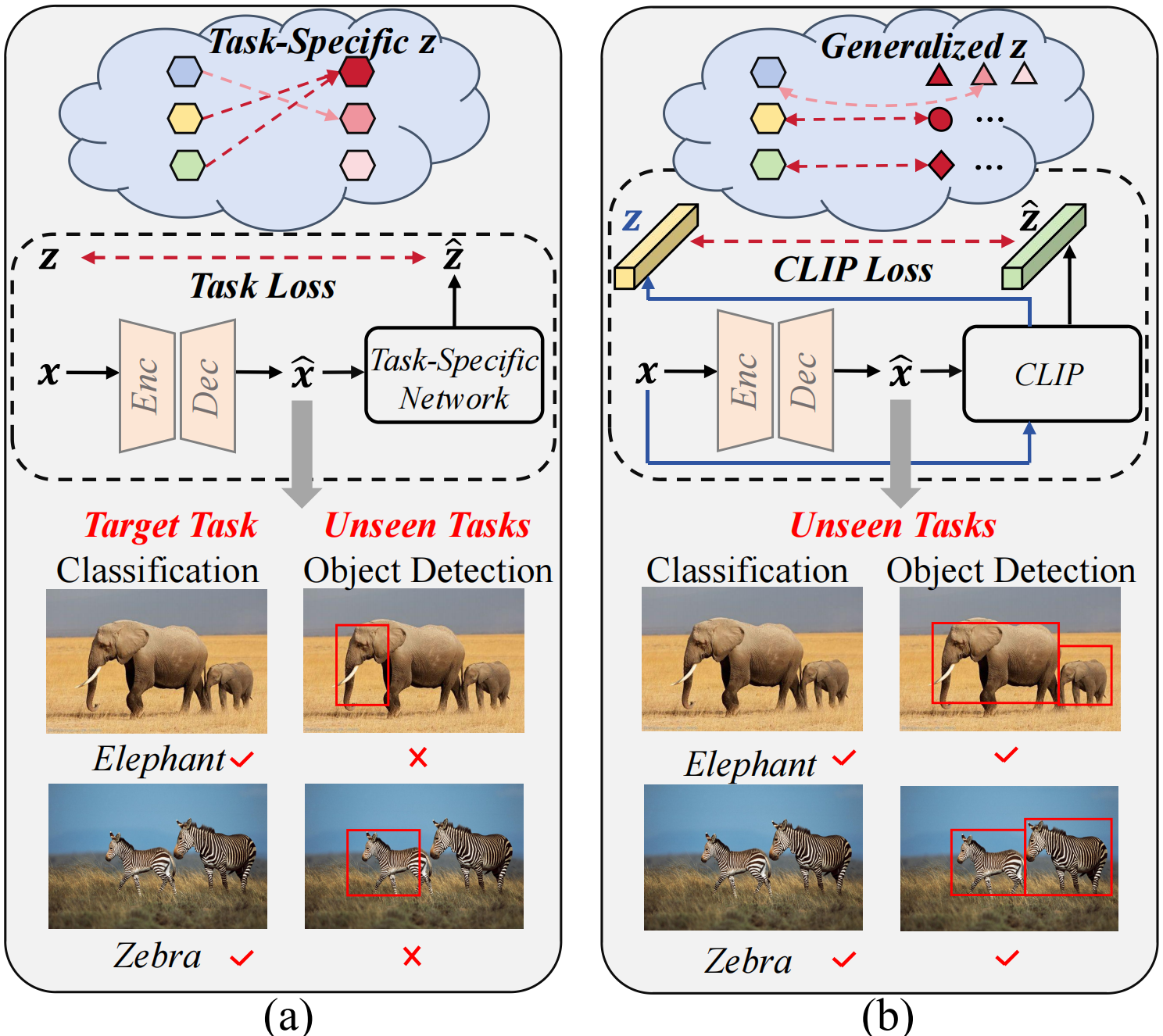}}
        \caption{(a) Most existing ICM models rely on task-specific supervision and fall short of generalization capacity in unseen scenarios.
        (b) 
        $z$ and $\hat{z}$ denote the latent representations satisfying downstream tasks from the original image $x$ and compressed image $\hat{x}$, respectively.}
     \label{fig:idea-highlight}
\end{figure}

To make up this issue for obtaining both human and machine-friendly images, Image Coding for Machine (ICM) is taken as a more promising direction with significant attention~\cite{10440522}.
%
Preliminary ICM explorations focus on conventional block-based codecs~\cite{sullivan2012overview, VVC}, making them aware of machine vision preferences by identifying semantic-related side information during preprocessing~\cite{huang2021visual, choi2020task}.
This information contributes to deriving appropriate block-wise coding parameters for analytics-friendly bit allocation.
In recent years, notable progress has been made on Learned Image Compression (LIC), providing the flexibility to integrate entire codec modules specifically tailored for machines.
Such codecs are jointly optimized under the supervision of downstream machine analytics models with respect to \textit{rate-perception-analytics} criteria~\cite{chamain2021end,li2022region}, which offers improved machine analytics performance.
However, the inherent gap between the representations of pixels and semantic features makes it challenging to achieve a good trade-off between perceptual quality and machine analytic performance~\cite{liu2019classification}.
To pursue enhanced machine analytics without sacrificing the perceptual quality, transfer-based methods~\cite{liu2022improving, chen2023transtic} have emerged as a new research trend, \textit{i.e.}, transferring human vision-oriented LIC to machine analytic purpose in a plug-and-play manner. 
However, these methods take separate bitstreams for different tasks, which falls short in terms of flexibility and overall compression efficiency.
Moreover, their generalization has long been overlooked and needs to be augmented, since existing methods are coupled with specific downstream machine analytics models for supervision.

%
%
%

Under these circumstances, our work explores to build a Unified and Generalized ICM (UG-ICM) paradigm aimed at satisfying both human perception and various machine analytics tasks (including unseen ones) with a unified bitstream.
%
\begin{figure*}[t]
      \centerline{\includegraphics[width=17cm]{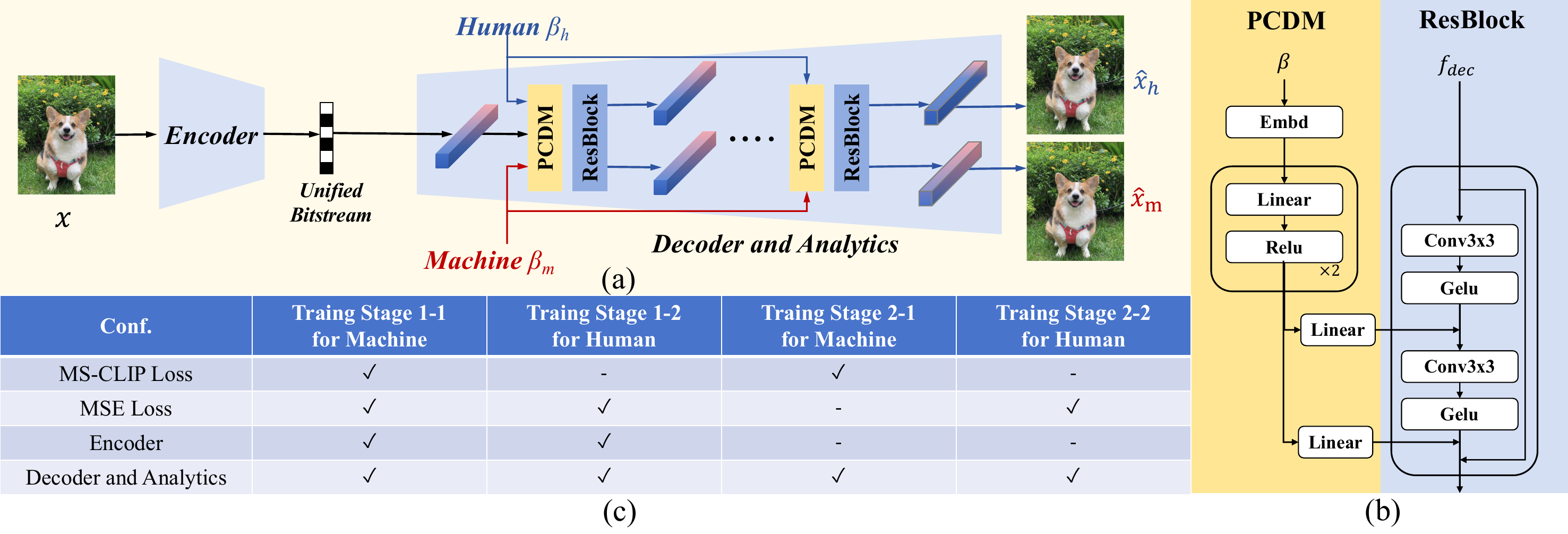}}
        \caption{
        (a) Compressing pipeline of the proposed UG-ICM.
        (b) Details of the proposed PCDM.
        (c) Modules and loss terms involved in the multi-stage training process.
        }
     \label{fig:framework}
\end{figure*}
For \textit{unification}, we introduce a conditional decoding strategy to turn such unified bitstream into different versions for human or machine preference by only adjusting a controllable parameter. 
Specifically, we develop a novel Preference Conditional Decoding Module (PCDM). which incorporates the controllable parameter into the decoder at different stages.
The PCDM produces a set of bias features based on the indexed human or machine preference, which are then incorporated into the image features to guide the decoding process.
Furthermore, we incorporate a multi-stage training strategy, for human perception and machine analytics alternately,  that stabilizes training and improves the training efficiency.

For \textit{generalization}, we construct a composite semantic supervision based on the pretraind Contrastive Language-Image Pre-training (CLIP) model~\cite{radford2021learning}, as illustrated in Fig.~\ref{fig:idea-highlight}.
%
Specifically, the feature similarity of compressed images and original ones is measured and regularized in the CLIP feature space with the cosine similarity as the generalized constraint that promotes machine vision analytics.
%
Moreover, such CLIP-based supervision covers hierarchical visual information from global-local-instance granularity, enhancing the capacity for capturing contextual information of different levels, termed Multi-Scale CLIP (MS-CLIP) loss.
It is noted that, the entire UG-ICM is trained in a self-supervised manner, without using any specific downstream model and annotations, but offering significant performance gains on a series of visual tasks.
We summarize our contributions as follows: 
\begin{itemize}
    \item We make the first exploration of leveraging the CLIP model to construct an intrinsic and generalized semantic constraint in the ICM field, pursuing strong generalization capacity in a self-supervised manner.
    \item We propose a unified ICM paradigm capable of providing satisfactory perceptual quality and improved analytics performance through a single bitstream.
    \item We introduce a preference conditional decoding module, enabling the decoding of a single bitstream into different versions, each specifically tailored to meet human or machine needs.
    \item Extensive experimental results have demonstrated that the proposed UG-ICM outperforms state-of-the-art ICM works in both generalized machine analytics and perceptual quality.
\end{itemize}

\section{Related Works}
\subsection{Learned Image Compression}
%
%
In the past decade, significant efforts have been devoted to Learned Image Compression (LIC) with considerable advancements. 
As a result, recent LIC developments have already shown superior performance compared to the state-of-the-art coding standard, VVC~\cite{VVC}.
In special,  Ball{\'e \textit{et al.} proposed the first end-to-end image compression network via the VAE architecture~\cite{balle2016end}, and subsequently introduced a hyper-prior~\cite{balle2018variational} and a local context-based~\cite{minnen2018joint} entropy model to enhance compression performance for \textit{rate-distortion}.
Along this vein, numerous efforts have been dedicated to further improving the entropy module. For instance, in ~\cite{guo2021causal}, Guo \textit{et al.} introduced global context information for enhanced entropy coding.
Meanwhile, checkerboard context models and channel-wise context were introduced to reduce entropy coding computational complexity in ~\cite{he2021checkerboard} and ~\cite{minnen2020channel}, respectively.
Moreover, by incorporating local spatial, global spatial, and channel contexts, Jiang \textit{et al.}~\cite{jiang2023mlic} proposed a multi-reference entropy model that achieved state-of-the-art performance.
Besides the entropy model, other works have adopted more powerful backbone networks, such as residual networks ~\cite{chen2022two}, invertible neural networks ~\cite{xie2021enhanced}, and Swin-Transformers ~\cite{liu2023learned}.

\subsection{Image Coding for Machine}
To simultaneously satisfy both human perception and machine analytics, preliminary Image Coding for Machine (ICM) works share a similar methodology that cascades the downstream machine analytics network to the output end of LIC. 
These are jointly trained with respect to the \textit{rate-distortion-analytics} criteria~\cite{chamain2021end, le2021image, wang2021end,liu2021semantics, torfason2018towards}.
However, as described in Liu \textit{et al.}~\cite{liu2019classification}, the inherent pixel-semantic gap inevitably leads this naive methodology to a trade-off scenario, \textit{i.e.}, improved machine analytics inevitably comes at the expense of perceptual quality.
Subsequently, some works explored the scalable coding strategy that initially transmits bottleneck features from downstream networks to fulfill machine analytics needs, low-level visual information is subsequently compressed for image reconstruction.
To this end, considerable effort has been invested in tailoring the framework for specific tasks~\cite{yang2021towards} and developing methods for decoupling semantic and low-level information~\cite{choi2022scalable, wang2021towards, liu2023icmh}.
Although these methods can fulfill machine analytics needs without sacrificing perceptual quality, the scalable strategies fall short of leveraging the correlation between high-level and low-level information, resulting in suboptimal compression rates.
Recently, transfer-based methods have emerged as a new research trend, \textit{i.e.}, transferring human vision-oriented LIC for machine analytic codecs in a plug-and-play manner \cite{liu2022improving, chen2023transtic, shen2024image}. 
However, these transfer-based methods typically involve a recompression process that produces separate bitstreams for human and machine needs, which lacks flexibility and efficiency in practical implementation.
%
%
%
%
%

\begin{figure}[t]
      \centerline{\includegraphics[width=9cm]{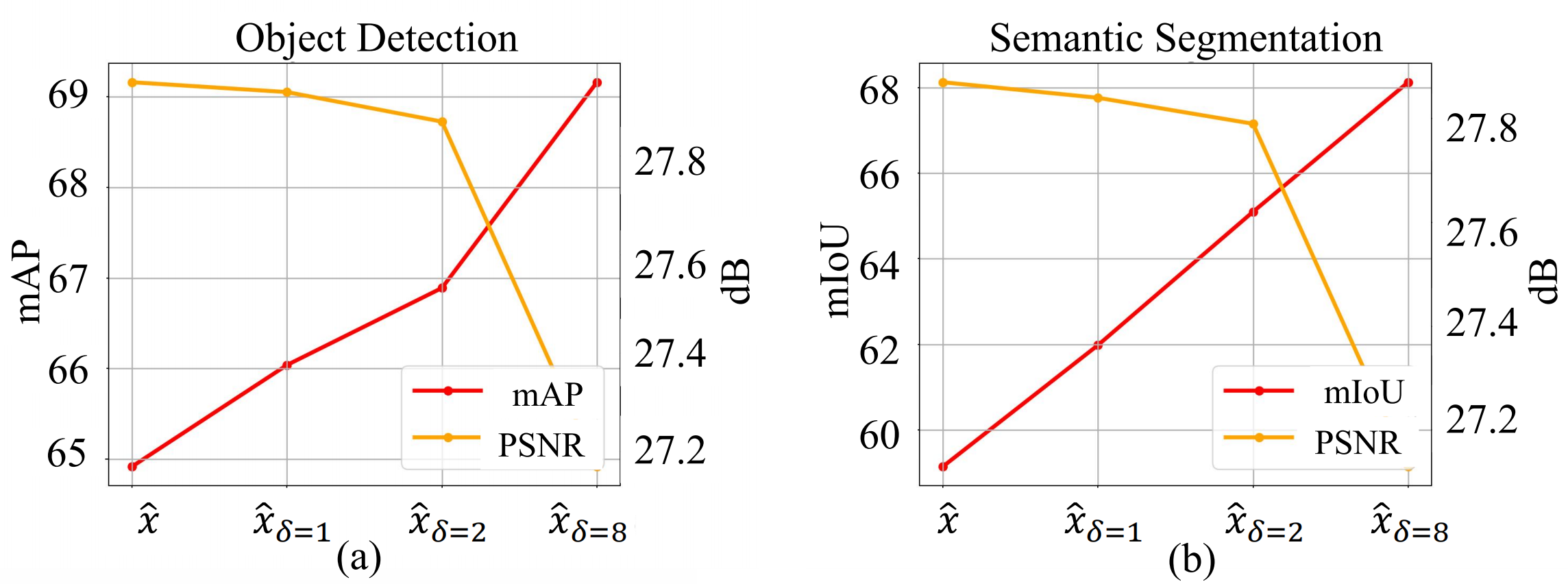}}
        \caption{
        Empirical analysis on generalization of taking CLIP as supervision constraint.}
     \label{fig:Attack_result}
\end{figure}
\section{Proposed Method}
\subsection{Overview and Motivations}
%
%
To address the neglected issues of existing ICM works, our work introduces a new paradigm, UG-ICM, characterized by its capacity to deliver satisfactory perceptual quality and generalized analytic performance with only a single bitstream.
The overall framework is illustrated in Fig.~\ref{fig:framework}.

Specifically, the input image $x$ would be first fed to an encoder $G(\cdot)$ to obtain the unified compact latent representation $\hat{y}$, containing the information required by both human perception and machine analytics,
\begin{equation}
\hat{y}= G(x).
\end{equation}
An entropy model $E(\cdot)$ can follow to estimate and constrain the entropy via $E( \hat{y} )$, which will be elaborated later.
Subsequently, the unified compact latent representation $\hat{y}$, along with the corresponding preference condition  $\beta\in\{\beta_{h}, \beta_{m}\}$ indicating human or machine preference, are jointly fed into the decoder $D(\cdot)$,
\begin{equation}
\hat{x}_{h}= D(\hat{y},\beta_{h}),\,\,\, \hat{x}_{m}=D(\hat{y},\beta_{m}),
\end{equation}
where $\hat{x}_{h}$ and $\hat{x}_{m}$ are reconstructed images favored by human perception and generalized machine analytics, respectively.

In our work, we aim for the model to own two key characteristics: first, $\hat{y}$ should be a \textit{general} representation that supports reconstruction for both human and machine vision; second, $\hat{y}$ can be \textit{generalized} to effectively handle a range of machine vision tasks.
To fulfill these goals, the preference-guided decoding process is realized by a Preference Conditional Decoding Module (PCDM), which generates preference-tailored features according to 
$\beta$ and biases the decoding process towards distinct versions.
To ensure the ICM fully recognizes the disparities between human and machine needs, we have specifically designed a multi-stage training strategy, with each stage supervised by human and machine-oriented criteria, respectively.
Moreover, to achieve strong generalization across various downstream tasks, we leverage the CLIP model to provide hierarchical machine-oriented supervision at glocal-local-instance levels, denoted as Multy-Scale (MS)-CLIP loss.
The proposed MS-CLIP loss, PCDM, and corresponding training strategy are detailed in the following subsections.
\subsection{Multi-Scale CLIP Supervision}
\subsubsection{Empirical Analysis on Generalization of Taking CLIP as Supervision Constraint.}
It is well known that the latent representation extracted by the CLIP model is capable of satisfying various unseen machine analysis tasks. 
However, in the context of ICM, the latent representation is required to own both compactness and pixel-level reconstruction capacity.
As such, we are motivated to first make a preliminary exploration of the generalization capacity of CLIP via leveraging it as the additional supervision for image construction rather than constraining latent representations.
To this end, we especially design an offline toy example to examine whether CLIP-based supervision is capable of facilitating generalized machine analytics meanwhile maintaining the perceptual quality.

In detail, denoting $\hat{x}$ as a reconstructed version of $x$ from a certain LIC, we conduct an offline image updating process under the supervision of CLIP,
\begin{align}
\hat{x}^{t+1} & \leftarrow \hat{x}^{t}-s \cdot sign[\nabla_{\hat{x}^{t}} (1-\cos(\text{CLIP}(\hat{x}^{t}),\text{CLIP}(x)],  \nonumber \\
\hat{x}^{t+1} & \leftarrow \hat{x} + \text{clamp}(\hat{x}^{t+1}-\hat{x}, - \delta,\delta),
\label{eq:attack}
\end{align}
where CLIP$(\cdot)$ refers to the image modal feature extractor, $\nabla$ signifies the corresponding gradient, $t$ denotes the updating iterations, $s$ represents the updating step size,  $\text{cos}(\cdot)$ measures the cosine similarity in CLIP feature space, and the $\delta$ is a pixel-wise constraint to ensure the perceptual quality.
%
Herein, we adopt 3,000 images from PASCAL VOC~\cite{everingham2010pascal} and compress them by MLIC++~\cite{jiang2023mlicpp}, updating them via Equ.~(\ref{eq:attack}) with a set of $\delta \in\{1,2,8\}$.
The updated images are subsequently fed into unseen object detection~\cite{redmon2018yolov3} and semantic segmentation~\cite{chen2018encoder} models to assess the machine analytics performance.

Corresponding results are shown in Fig.~\ref{fig:Attack_result}, where improvements in various machine analytics tasks can be easily observed with also favorable perceptual quality. 
These results demonstrate that CLIP is also capable of providing strong generalization capacity from the perspective of image reconstruction guidance, which has motivated us to further study and incorporate it into the ICM framework.
\begin{figure}[t]
      \centerline{\includegraphics[width=8.5cm]{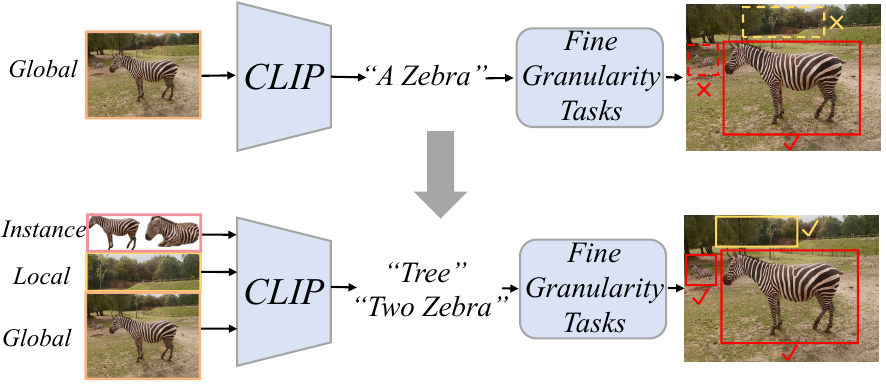}}
        \caption{Illustration of the proposed Multi-Scale CLIP loss.}
     \label{fig:CLIP Supervision}

\end{figure}
\subsubsection{Multi-Scale CLIP Loss.}
Although the strong generalization capacity of CLIP has been demonstrated, there is still an issue that CLIP itself is an image-level modeling tool and trained to match an image as a whole to a text description.
It is difficult to provide region- or pixel-level fine-grained information and may overlook the contextual information of less salient regions~\cite{zhong2022regionclip,
gao2022pyramidclip}.
To construct a comprehensive semantic constraint, we propose a novel MS-CLIP loss that applies CLIP supervision at global, local, and instance levels, as shown in Fig.~\ref{fig:CLIP Supervision}.

To comprehensively capture the local contextual information from the input image, we 
constraint the perspective field of the CLIP model to the instance level.
In special, an instance segmentation model~\cite{wang2022freesolo} is leveraged to obtain a set of instance-wise pairs  $[i_1,i_2,...,i_k]$ and $[\hat{i_1},\hat{i_2},...,\hat{i_k}]$ from the input image $x$ and the reconstructed image $\hat{x}$.
%
%
%
The cosine similarity between them would be calculated,
\begin{equation}
\mathcal{L}_{ins}(x,\hat{x})=\sum_{j=1}^{k} (1 - \cos(\text{CLIP}(i_j), \text{CLIP}(\hat{i_j}))).
\end{equation}
%

%
Moreover, to obtain more fine-grained information for potentially more generalized semantic clues to capture relationships among different models, 
we incorporate a local-wise supervision method to enhance generalization capabilities across various local perspectives.
In special, during the training process, we random crop a local-wise patch pairs $[l,\hat{l}]$ from the input $x$ and reconstructed version $\hat{x}$, and calculated their cosine similarity at the CLIP feature space,
\begin{equation}
\mathcal{L}_{loc}(x,\hat{x})= 1 - \cos(\text{CLIP}(l), \text{CLIP}(\hat{l})).
\end{equation}

Thus, by incorporating global-wise CLIP-based supervision $\mathcal{L}_{glo}$,
\begin{equation}
\mathcal{L}_{glo}(x,\hat{x})= 1 - \cos(\text{CLIP}(x), \text{CLIP}(\hat{x})),
\end{equation}
the established multi-scale CLIP loss $\mathcal{L}_{MC}$ is expected to provide a comprehensive understanding of the semantic divergence between input image pairs, which can be formulated by,
\begin{equation}
\mathcal{L}_{MC}(x,\hat{x})= \mathcal{L}_{glo}(x,\hat{x})+\mathcal{L}_{loc}(x,\hat{x})+\mathcal{L}_{ins}(x,\hat{x}).
\end{equation}

\subsection{Preference Conditional Decoding Module}
The decoder of the proposed UG-ICM is responsible for generating different versions of reconstructed images tailored to human or machine needs from a unified bitstream. 
This capability is enabled by a specially designed Preference Conditional Decoding Module (PCDM), which is integrated into each decoding block.
As illustrated in Fig.~\ref{fig:framework} (b), during the decoding process, the preference condition is indicated by $\beta \in \left\{ \beta_h,\beta_m \right\}$ for human or machine needs respectively. 
The preference condition is subsequently fed into a 2-layer MLP, yielding a corresponding preference conditional feature, 
$f_{\beta}$. 
This feature, $f_{\beta}$, is then fused with the image feature to bias the decoding process towards its indicated preference,
\begin{equation}
f_{PCDM}=f_{\beta}W_{d}+f_{dec},
\end{equation}
where  $f_{PCDM}$ and $f_{dec}$ denote the output and the input features of the PCDM module respectively,  $W_d$ is the feature element-wise weighting map with $d$ denoting the channel depth aligned with the input features.
In coding practice, the $[\beta_h, \beta_m]$ are valued at $[0,1]$.

%
%
%
%
%
%
\begin{figure*}[t]
\centerline{\includegraphics[width=17cm]{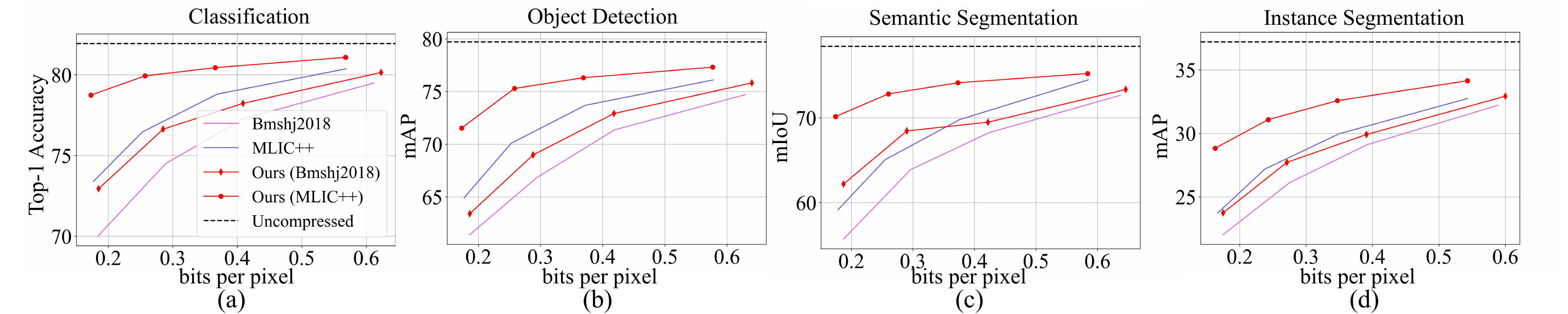}}
        \caption{Machine analytics performance comparisons between the proposed UG-ICM and the backbone compression networks.}
     \label{fig:Machine viosn}
\end{figure*}
\begin{figure*}[t]
      \centerline{\includegraphics[width=17cm]{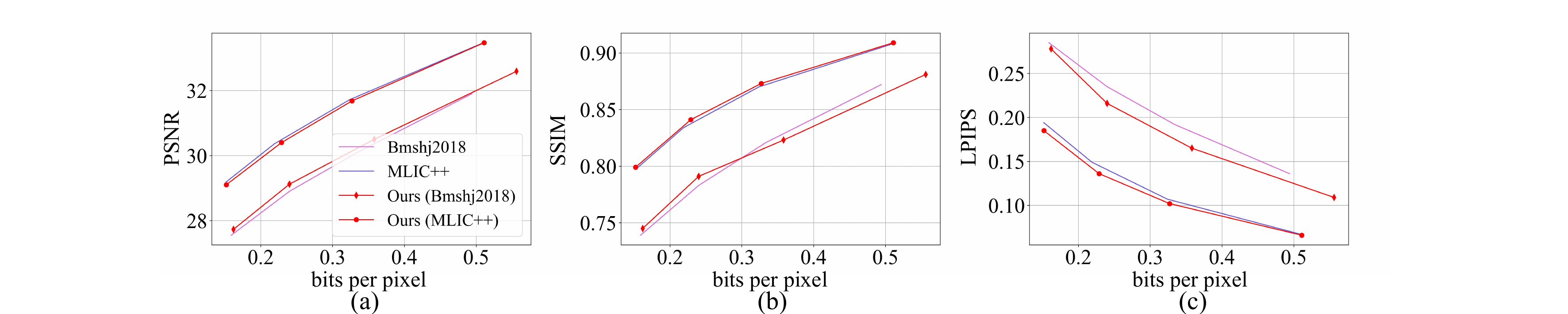}}
        \caption{Perceptual quality comparisons between the proposed UG-ICM and the backbone compression networks.}
     \label{fig:Human viosn}
\end{figure*}
\subsection{Training and Supervision}
We develop a two-stage training strategy as shown in Fig.~\ref{fig:framework}~(c) to achieve optimal performance in two distinct objectives within the proposed UG-ICM framework.
In detail, the first stage optimizes the entire framework with the goal of fully extracting visual information for both human perception and machine analytics. The second stage specifically emphasizes the conditional decoder to further improve the performance with respect to two distinct objectives. 
%
%
\subsubsection{First Stage.}
As for the first stage, the encoder $G_{\theta}$, entropy model $E_{\phi}$, and the conditional decoder $D_{\psi}$, parameterized by $\theta$, $\phi$, $\psi$, respectively, are jointly optimized by alternating between \textit{rate-perception} and \textit{rate-analytics} criteria.
%
%
In particular, we split the training iterations within one epoch into two sessions, and the corresponding training process can be formulated by:
\begin{align}
\textit{S 1-1:}\bar{\theta}, \bar{\phi}, \bar{\psi} &= \mathop{\arg\min}\limits_{(\phi,\psi,\theta)}  
\sum\limits_{(x\in X)}  \lambda \cdot \mathcal{L}_{2}(x, D_{\psi}({G_{\theta}(x)},\beta_h)) \notag \\
&\quad + E_{\phi}(\hat{y}), \nonumber
\end{align}
\begin{align}
\textit{S 1-2:}\Tilde{\theta}, \Tilde{\phi}, \Tilde{\psi} &= \mathop{\arg\min}\limits_{(\bar{\theta}, \bar{\phi}, \bar{\psi})} 
\sum\limits_{(x\in X)}  \lambda \cdot\big(  \mathcal{L}_{2}(x, D_{\bar{\psi}}({G_{\bar{\theta}}(x)},\beta_m) \notag \\
&\quad +  \mathcal{L}_{MC}(x, D_{\bar{\psi}}({G_{\bar{\theta}}(x)},\beta_m)\big)+ E_{\bar{\phi}}(\hat{y}), \nonumber
\end{align}
where $\lambda$ is a Lagrange parameter to obtain the ICM models at different compression levels.
%
\subsubsection{Second Stage.}
In the second stage, the parameters of the encoder and entropy model are frozen, thereby focusing optimization efforts solely on the decoder.
Moreover, the iterations are half split into two sessions for each epoch, the training processes are formulated by,
\begin{equation} 
\textit{S 2-1:}\bar{\psi}=\mathop{\arg\min}\limits_{(\psi)}\sum\limits_{(x\in X)} \mathcal{L}_{2}(x, D_{\psi}({G_{\theta}(x)},\beta_h)), \nonumber
\end{equation}
\begin{equation}
\textit{S 2-2:}\Tilde{\psi}=\mathop{\arg\min}\limits_{( \bar{\psi})}\sum\limits_{(x\in X)} \mathcal{L}_{MC}(x, D_{\bar{\psi}}({G_{\theta}(x)},\beta_m)). \nonumber
\end{equation}

\begin{figure*}[t]
      \centerline{\includegraphics[width=17cm]{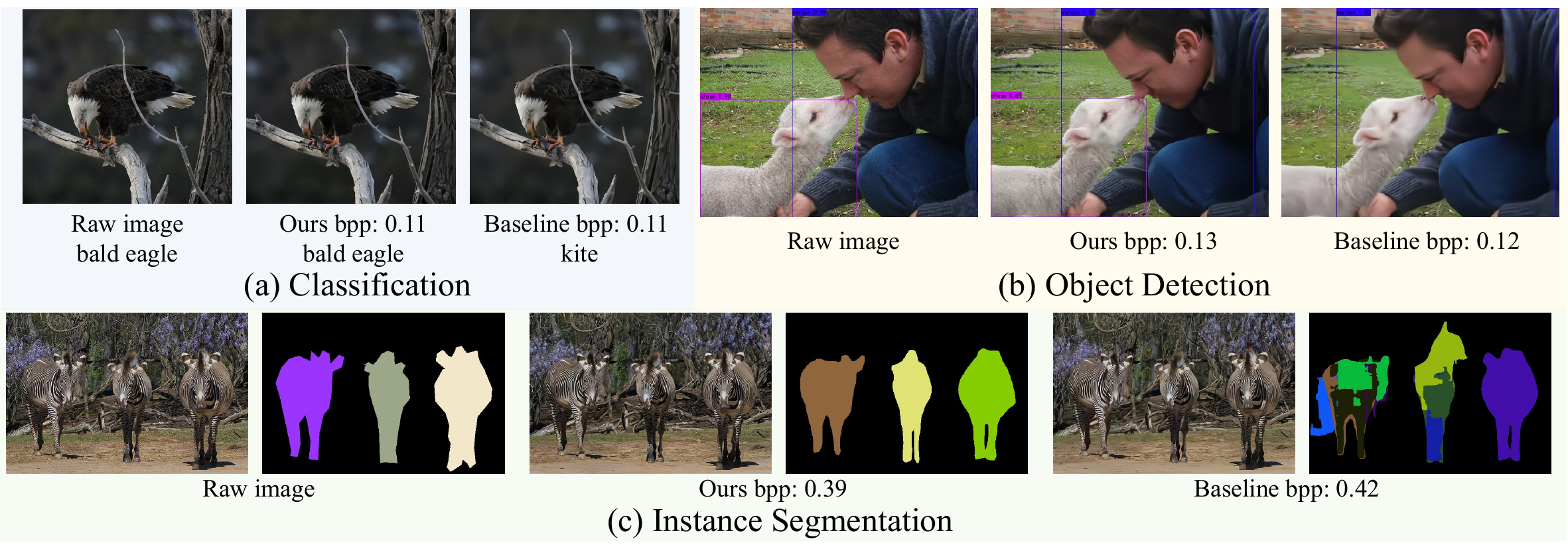}}
        \caption{Visualization of the machine analytics performance regarding      
        classification (a), object detection (b), and instance segmentation (c), where the `Baseline' denotes the backbone compression network MLIC++.}
     \label{fig:result show}
\end{figure*}
\section{Experiments}
We first implement the proposed UG-ICM on different backbone LIC networks, examining its effectiveness by comparing the LIC networks with and without our implementations.
Subsequently, we compare the UG-ICM with cutting-edge ICM methods in terms of perceptual quality and machine analytics performance improvement across a variety of downstream tasks.
Moreover, comprehensive ablation studies have been conducted to demonstrate the effectiveness of the proposed PCDM and  Multi-Scale CLIP Loss.

\subsection{Experimental Settings}
\label{ssec:eval_aafgda}
\subsubsection{Benchmark.} To comprehensively examine the generalized machine analytics performance of the proposed UG-ICM, four commonly employed downstream tasks are employed, including classification, object detection, instance segmentation, and semantic segmentation, all of which are unseen by our method. 
Additionally, perceptual quality is thoroughly assessed using various indices.
\begin{itemize}
    \item \textbf{Classification:} 25,000 images from the ILSVRC 2012 dataset~\cite{deng2009imagenet} (ImageNet-1k), spanning 1,000 categories, are employed for testing.
    ResNet101~\cite{he2016deep} well-trained on ImageNet is employed for assessment.
    The performance is indexed by Top-1 accuracy.
    \item \textbf{Object Detection:}  2,150 images from the testing set of PASCAL VOC~\cite{everingham2010pascal} are employed for testing. The YOLOv3~\cite{redmon2018yolov3} well-trained by VOC is used for assessment.
    The performance is indexed by mAP for IoU = 0.5. 
    \item \textbf{Instance Segmentation:} The subset \textit{val2017} of COCO dataset~\cite{lin2014microsoft} are employed for testing. 
    The  Mask-RCNN~\cite{he2017mask}  well-trained by COCO is employed for assessment.
    Corresponding performance is indexed by mAP for IoU $\in$ $[0.5: 0.05: 0.95] $.
     \item \textbf{Semantic Segmentation:} 
   1,450 images from PASCAL VOC are employed for testing.
   DeepLabv3+~\cite{chen2018encoder}  well-trained by VOC is employed for assessment.
      The performance is indexed by mIoU.
   
    \item \textbf{Human Perception:} The Kodak24 dataset is employed to assess the perceptual quality using three full-reference image quality assessment measures, \textit{i.e.}, PSNR, SSIM, and LPIPS~\cite{zhang2018unreasonable}.
\end{itemize}
\subsubsection{Anchors.} We implement the proposed UG-ICM on two LIC networks, including the milestone \textit{Bmshj2018}~\cite{balle2018variational} and the cutting-edge MLIC++~\cite{jiang2023mlicpp}.
To ensure fairness, the LIC with and without the implementation of UG-ICM are both trained from scratch, with the same training set. 

As for the compared ICM works, we employ three works with different methodologies and pipelines for a comprehensive examination. These include two supervised approaches: \textit{Chamain's}~\cite{chamain2021end} and TransTIC~\cite{chen2023transtic}, along with a self-supervised approach, SA-ICM~\cite{shindo2024image}.
In detail,
\begin{itemize}
    \item \textit{Chamain's}: represents the conventional ICM paradigm, which straightforwardly adopts the machine analytics performance as an extra loss term for network optimization.
    We implement it on the MLIC++ backbone and retrain it with corresponding downstream models to ensure fair comparisons.
    \item TransTIC: is the cutting-edge transfer-based ICM, adapting the LIC to machine analytics needs in a plug-and-play manner.
    It should be noted that, with this method, images need to be recompressed into another bitstream when adapting to a new task.
    \item SA-ICM: is a self-supervised ICM approach that adopts the segmentation map from SAM~\cite{kirillov2023segment} to identify regions of interest, which are then allocated more coding resources during the compression process.

\end{itemize}

\subsubsection{Implementation Details.}
80,000 images from the subset \textit{train2017} of COCO are employed for tranining.
The whole network is implemented in Pytorch with CUDA support and is trained on a single NVIDIA A800-80G GPU.
The CLIP (ViT-B/32) is adopted for our MS-CLIP implementation.
In the first stage of the training process, we use the Adam optimizer with a learning rate of 1e-4 and train for 200 epochs. For the second stage, we reduce the learning rate to 1e-5 and train for an additional 10 epochs.
The UG-ICM and corresponding LIC backbone networks are each trained four times with different Lagrange parameters, achieving four distinct compression rates regarding bits-per-pixel (bpp).
\begin{figure*}[t]
      \centerline{\includegraphics[width=17cm]{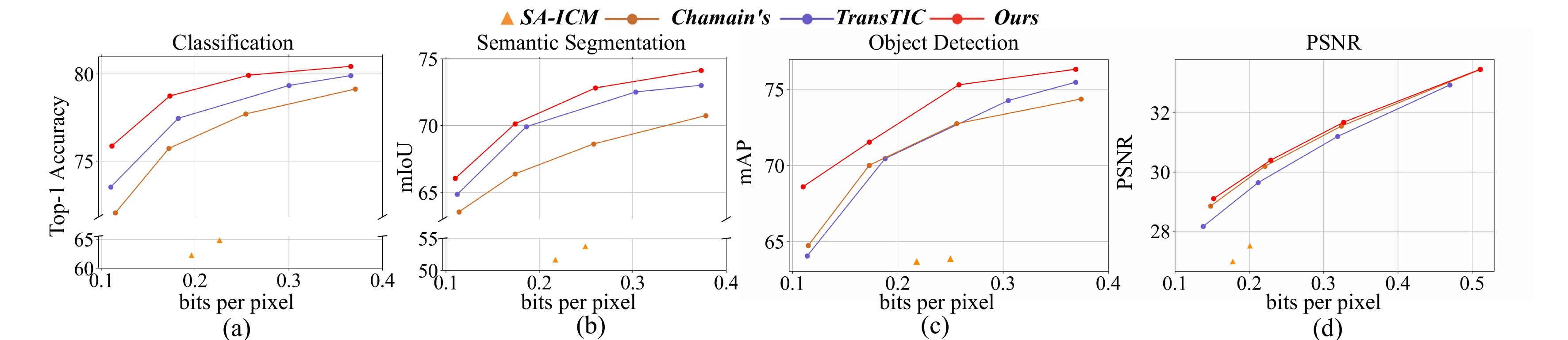}}
        \caption{Performance comparison between the proposed UG-ICM and employed anchors regarding classification (a), semantic segmentation (b), objection detection (c), and perceptual quality indexed by PSNR (d).}
     \label{fig:Other_ICM_Methods}
\end{figure*}
\begin{figure}[t]
      \centerline{\includegraphics[width=8.5cm]{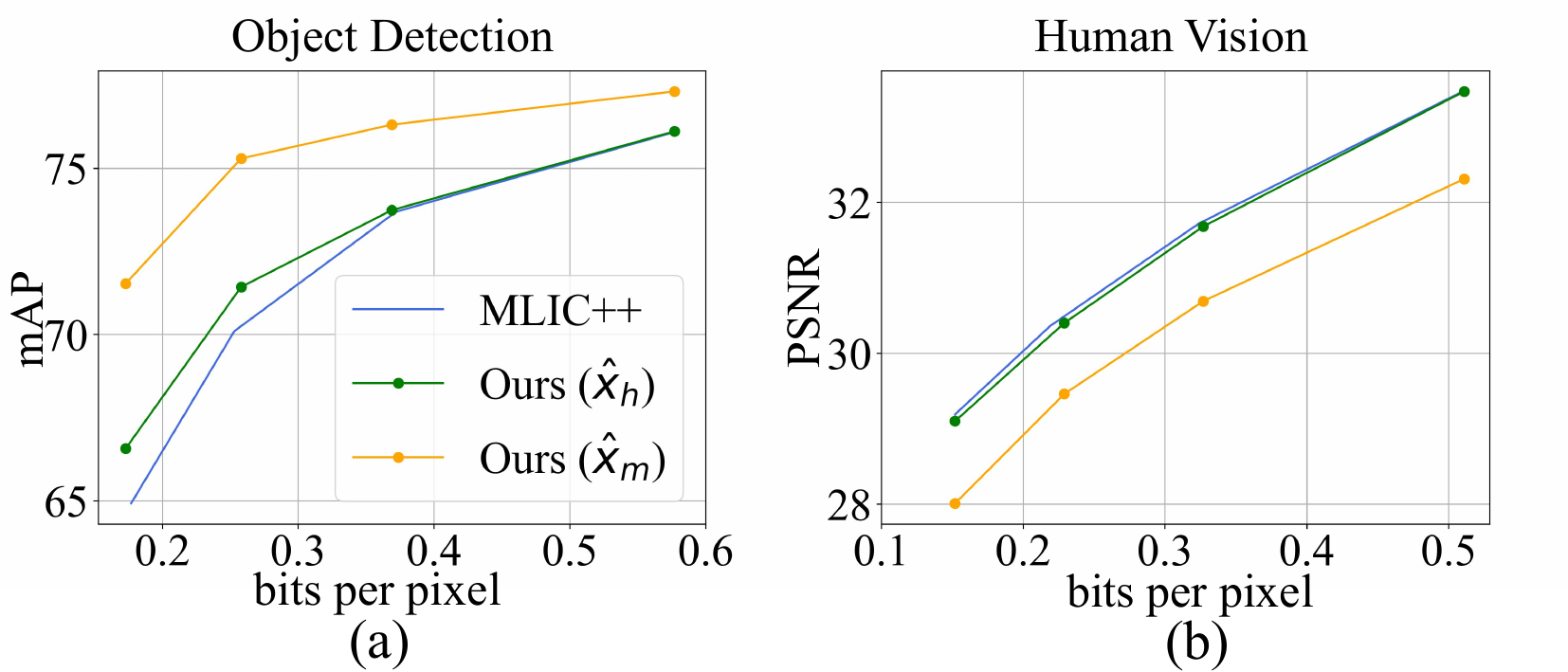}}
        \caption{Performance comparisons between decoded images tailored for human and machine preferences.}
     \label{fig:ablation1}
\end{figure}

\subsection{Experimental Results}
\subsubsection{Comparison on Different Backbones.} As mentioned above, the proposed UG-ICM is implemented on two LIC backbones, Bmshji2018 and MLIC++.
In this subsection, we compare the UG-ICM with the vanilla backbone in terms of perceptual quality and various unseen downstream models, demonstrating the effectiveness of our implementation.
The comparison results for machine analytics are presented in Fig.~\ref{fig:Machine viosn}. 
Encouraging observations have been made, showing that the proposed method brings overall improvement across multiple downstream tasks and bitrates.

Specifically, with the Bmshj backbone, the proposed method achieves an average analytics gain of $1.7\%$ in top-1 accuracy, $1.7\%$ in mAP, $3.2\%$ in mIOU, and $1.2\%$ in mAP under the same bits-per-pixel (bpp) level for classification, object detection, semantic segmentation, and instance segmentation, respectively.
As for the  MLIC++ backbone, the improvement becomes even larger, with an average bpp-analytics gain of $2.8\%$, $3.9\%$, $6.0\%$, and $3.2\%$ regarding the four downstream tasks.
Considering that both the downstream networks and corresponding annotations are unseen by our UG-ICM, the observed improvements are significant, demonstrating that the rich image modal knowledge from CLIP is comprehensively leveraged by our scheme.
To gain intuition into the analytics performance improvement, a group of examples is provided in Fig.~\ref{fig:result show}.

Comparison results in terms of PSNR, SSIM, and LPIPS are provided in Fig.~\ref{fig:Human viosn} (a), (b), and (c) respectively. 
As shown, the human-preference version of our UG-ICM maintains comparable perceptual quality to the corresponding backbones, owing to the proposed PCDM module and the multi-stage training strategy that sensitively aligns the compression module with both human and machine needs.
%

%
%
\subsubsection{Comparison to State-of-the-art ICM Methods.}
This subsection compares our UG-ICM (MLIC++) with other ICM methods, covering both machine vision and human vision aspects.
Comparison results are provided in Fig.~\ref{fig:Other_ICM_Methods}, demonstrating the superiority of our proposed UG-ICM against others regarding both machine analytics and perceptual quality.
In particular, compared with the conventional ICM framework \textit{Chamain's}, which adopts the same compression backbone as ours, the naive method of incorporating machine analytics preferences and the empirically decided bits allocation strategy fall short in achieving a good balance between perceptual quality and multiple downstream tasks.
Moreover, compared with TransTIC, the proposed UG-ICM still possesses significant advantages in machine analytics. This is because TransTIC employs a \textit{plug-and-play} strategy, which avoids changing the parameters of the compression backbone, ultimately leading to limited analytics improvement.
\subsection{Ablation Studies}
\subsubsection{Effectiveness Examination of the PCDM.} The proposed UG-ICM is capable of decoding a single bitstream into different versions, facilitated by the incorporated PCDM module.
Herein, we assess its effectiveness by comparing the decoded versions tailored for human preference, $\hat{x}_{h}$,  machine preference, $\hat{x}_{m}$, as well as the decoded images from the compression backbone.
As shown in Fig.~\ref{fig:ablation1}, $\hat{x}_{m}$ and $\hat{x}_{h}$ exhibit a clear bias towards their corresponding preferences. Specifically, $\hat{x}_{m}$ achieves an average gain of $3\%$ in bpp-mAP compared to $\hat{x}_{h}$ regarding object detection, but falls short by 1dB bpp-PSNR concerning human perception.
\begin{figure}[t]
      \centerline{\includegraphics[width=8.5cm]{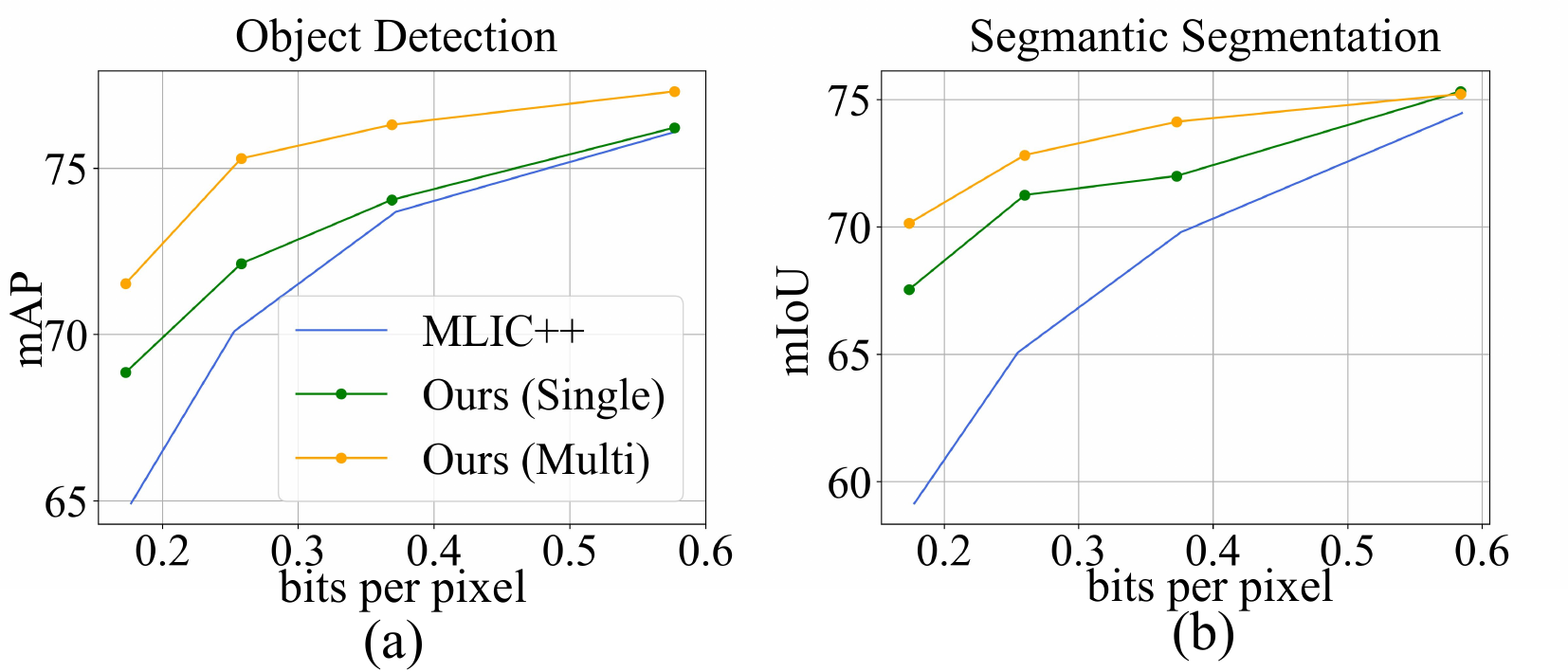}}
        \caption{Ablation studies of Multi-Scale CLIP Loss.}
     \label{fig:ablation2}
\end{figure}
\subsubsection{Ablation Studies of The Multi-Scale CLIP Loss.}
We ablate the multi-scale strategy from the CLIP-based supervision by retaining the whole framework with global-wise CLIP loss.
Comparison results between our full version `Ours (Multi)' and the ablated version `Ours (Single)' regarding object detection and semantic segmentation are provided in Fig.~\ref{fig:ablation2}.
As shown, the multi-scale supervision strategy significantly improves machine analytics by enhancing the ability to capture the semantic information of small objects, which are often overlooked. In contrast, vanilla CLIP models tend to prioritize capturing global context.

\section{Conclusion}
This paper proposes a novel Unified and Generalized (UG-ICM).
It is capable of providing satisfactory perceptual quality and improved machine analytics performance across various tasks with a single bitstream.
Specifically, we first incorporate a conditional decoding module that introduces human or machine needs as extra conditions, enabling the bitstream to be decoded into different versions for corresponding preferences.
Furthermore, we explore the generalization capacity of the CLIP model and propose global-to-instance-wise CLIP supervision, accommodating preferences for different scale semantic information across various machine analytics tasks.
Experimental results and ablation studies have demonstrated the effectiveness of our design.

\section{Acknowledgments}
This work was in part by the Basic and Frontier Research Project of PCL, the Major Key Project of PCL, in part by Guangdong Basic and Applied Basic Research Foundation under Grant 2024A1515010454, in part by the Open Research Fund from Guangdong Laboratory of Artificial Intelligence and Digital Economy (SZ) under Grant No. GML-KF-24-27, in part by the Natural Science Foundation of Guangdong Province under Grant 2023A1515011667, in part by the Science and Technology Major Project of Shenzhen under Grant 202302D074, in part by the Key Basic Research Foundation of Shenzhen under Grant JCYJ20220818100205012, and in part by the Guangdong Basic and Applied Basic Research Foundation under Grant 2023B1515120020.

\appendix

\bibliography{aaai25}

\begin{thebibliography}{46}
\providecommand{\natexlab}[1]{#1}

\bibitem[{Ball{\'e}, Laparra, and Simoncelli(2017)}]{balle2016end}
Ball{\'e}, J.; Laparra, V.; and Simoncelli, E.~P. 2017.
\newblock End-to-end optimized image compression.
\newblock In \emph{Proceedings of the International Conference on Learning Representations}.

\bibitem[{Ball{\'e} et~al.(2018)Ball{\'e}, Minnen, Singh, Hwang, and Johnston}]{balle2018variational}
Ball{\'e}, J.; Minnen, D.; Singh, S.; Hwang, S.~J.; and Johnston, N. 2018.
\newblock Variational image compression with a scale hyperprior.
\newblock In \emph{Proceedings of the International Conference on Learning Representations}.

\bibitem[{Bross et~al.(2021)Bross, Wang, Ye, Liu, Chen, Sullivan, and Ohm}]{VVC}
Bross, B.; Wang, Y.; Ye, Y.; Liu, S.; Chen, J.; Sullivan, G.; and Ohm, J. 2021.
\newblock Overview of the Versatile Video Coding ({VVC}) Standard and its Applications.
\newblock \emph{IEEE Transactions on Circuits and Systems for Video Technology}, 31(10): 3736--3764.

\bibitem[{Chamain et~al.(2021)Chamain, Racap{\'e}, B{\'e}gaint, Pushparaja, and Feltman}]{chamain2021end}
Chamain, L.~D.; Racap{\'e}, F.; B{\'e}gaint, J.; Pushparaja, A.; and Feltman, S. 2021.
\newblock End-to-end optimized image compression for machines, a study.
\newblock In \emph{Proceedings of the Data Compression Conference}, 163--172.

\bibitem[{Chen, Xu, and Wang(2022)}]{chen2022two}
Chen, F.; Xu, Y.; and Wang, L. 2022.
\newblock Two-stage octave residual network for end-to-end image compression.
\newblock In \emph{Proceedings of the AAAI Conference on Artificial Intelligence}, volume~36, 3922--3929.

\bibitem[{Chen et~al.(2018)Chen, Zhu, Papandreou, Schroff, and Adam}]{chen2018encoder}
Chen, L.-C.; Zhu, Y.; Papandreou, G.; Schroff, F.; and Adam, H. 2018.
\newblock Encoder-decoder with atrous separable convolution for semantic image segmentation.
\newblock In \emph{Proceedings of the European Conference on Computer Vision}, 801--818.

\bibitem[{Chen et~al.(2023)Chen, Weng, Kao, Chien, Chiu, and Peng}]{chen2023transtic}
Chen, Y.-H.; Weng, Y.-C.; Kao, C.-H.; Chien, C.; Chiu, W.-C.; and Peng, W.-H. 2023.
\newblock Transtic: Transferring transformer-based image compression from human perception to machine perception.
\newblock In \emph{Proceedings of the IEEE International Conference on Computer Vision}, 23297--23307.

\bibitem[{Chen et~al.(2020)Chen, Fan, Wang, Duan, Lin, and Kot}]{chenzhuo}
Chen, Z.; Fan, K.; Wang, S.; Duan, L.; Lin, W.; and Kot, A. 2020.
\newblock Toward Intelligent Sensing: Intermediate Deep Feature Compression.
\newblock \emph{IEEE Transactions on Image Processing}, 29: 2230--2243.

\bibitem[{Chen et~al.(2019)Chen, Fan, Wang, Duan, Lin, and Alex}]{chen2019lossy}
Chen, Z.; Fan, K.; Wang, S.; Duan, Y.; Lin, W.; and Alex, K. 2019.
\newblock Lossy intermediate deep learning feature compression and evaluation.
\newblock In \emph{Proceedings of the ACM International Conference on Multimedia}, 2414--2422.

\bibitem[{Choi and Baji{\'c}(2022)}]{choi2022scalable}
Choi, H.; and Baji{\'c}, I.~V. 2022.
\newblock Scalable image coding for humans and machines.
\newblock \emph{IEEE Transactions on Image Processing}, 31: 2739--2754.

\bibitem[{Choi and Han(2020)}]{choi2020task}
Choi, J.; and Han, B. 2020.
\newblock Task-aware quantization network for jpeg image compression.
\newblock In \emph{Proceedings of European Conference on Computer Vision}, 309--324.

\bibitem[{Deng et~al.(2009)Deng, Dong, Socher, Li, Li, and Fei-Fei}]{deng2009imagenet}
Deng, J.; Dong, W.; Socher, R.; Li, L.-J.; Li, K.; and Fei-Fei, L. 2009.
\newblock Imagenet: A large-scale hierarchical image database.
\newblock In \emph{Proceedings of the IEEE Conference on Computer Vision and Pattern Recognition}, 248--255.

\bibitem[{Everingham et~al.(2010)Everingham, Van~Gool, Williams, Winn, and Zisserman}]{everingham2010pascal}
Everingham, M.; Van~Gool, L.; Williams, C.~K.; Winn, J.; and Zisserman, A. 2010.
\newblock The pascal visual object classes (voc) challenge.
\newblock \emph{International Journal of Computer Vision}, 88: 303--338.

\bibitem[{Gao et~al.(2022)Gao, Liu, Xu, Zhang, Li, Ji, and Shen}]{gao2022pyramidclip}
Gao, Y.; Liu, J.; Xu, Z.; Zhang, J.; Li, K.; Ji, R.; and Shen, C. 2022.
\newblock Pyramidclip: Hierarchical feature alignment for vision-language model pretraining.
\newblock \emph{Neural Information Processing Systems}, 35: 35959--35970.

\bibitem[{Guo et~al.(2021)Guo, Zhang, Feng, and Chen}]{guo2021causal}
Guo, Z.; Zhang, Z.; Feng, R.; and Chen, Z. 2021.
\newblock Causal contextual prediction for learned image compression.
\newblock \emph{IEEE Transactions on Circuits and Systems for Video Technology}, 32(4): 2329--2341.

\bibitem[{He et~al.(2021)He, Zheng, Sun, Wang, and Qin}]{he2021checkerboard}
He, D.; Zheng, Y.; Sun, B.; Wang, Y.; and Qin, H. 2021.
\newblock Checkerboard context model for efficient learned image compression.
\newblock In \emph{Proceedings of the IEEE Conference on Computer Vision and Pattern Recognition}, 14771--14780.

\bibitem[{He et~al.(2017)He, Gkioxari, Doll{\'a}r, and Girshick}]{he2017mask}
He, K.; Gkioxari, G.; Doll{\'a}r, P.; and Girshick, R. 2017.
\newblock Mask r-cnn.
\newblock In \emph{Proceedings of the IEEE International Conference on Computer Vision}, 2961--2969.

\bibitem[{He et~al.(2016)He, Zhang, Ren, and Sun}]{he2016deep}
He, K.; Zhang, X.; Ren, S.; and Sun, J. 2016.
\newblock Deep residual learning for image recognition.
\newblock In \emph{Proceedings of the IEEE Conference on Computer Vision and Pattern Recognition}, 770--778.

\bibitem[{Huang et~al.(2021)Huang, Jia, Wang, and Ma}]{huang2021visual}
Huang, Z.; Jia, C.; Wang, S.; and Ma, S. 2021.
\newblock Visual analysis motivated rate-distortion model for image coding.
\newblock In \emph{Proceedings of IEEE International Conference on Multimedia and Expo}, 1--6.

\bibitem[{Jiang and Wang(2023)}]{jiang2023mlicpp}
Jiang, W.; and Wang, R. 2023.
\newblock MLIC++: Linear Complexity Multi-Reference Entropy Modeling for Learned Image Compression.
\newblock In \emph{ICML Workshop Neural Compression: From Information Theory to Applications}.

\bibitem[{Jiang et~al.(2023)Jiang, Yang, Zhai, Ning, Gao, and Wang}]{jiang2023mlic}
Jiang, W.; Yang, J.; Zhai, Y.; Ning, P.; Gao, F.; and Wang, R. 2023.
\newblock Mlic: Multi-reference entropy model for learned image compression.
\newblock In \emph{Proceedings of the ACM International Conference on Multimedia}, 7618--7627.

\bibitem[{Kirillov et~al.(2023)Kirillov, Mintun, Ravi, Mao, Rolland, Gustafson, Xiao, Whitehead, Berg, Lo et~al.}]{kirillov2023segment}
Kirillov, A.; Mintun, E.; Ravi, N.; Mao, H.; Rolland, C.; Gustafson, L.; Xiao, T.; Whitehead, S.; Berg, A.~C.; Lo, W.-Y.; et~al. 2023.
\newblock Segment anything.
\newblock In \emph{Proceedings of the IEEE/CVF International Conference on Computer Vision}, 4015--4026.

\bibitem[{Le et~al.(2021)Le, Zhang, Cricri, Ghaznavi-Youvalari, and Rahtu}]{le2021image}
Le, N.; Zhang, H.; Cricri, F.; Ghaznavi-Youvalari, R.; and Rahtu, E. 2021.
\newblock Image coding for machines: an end-to-end learned approach.
\newblock In \emph{Proceedings of the IEEE International Conference on Acoustics, Speech and Signal Processing}, 1590--1594.

\bibitem[{Li et~al.(2022)Li, Ye, Liang, Wang, and Han}]{li2022region}
Li, B.; Ye, L.; Liang, J.; Wang, Y.; and Han, J. 2022.
\newblock Region-of-interest and channel attention-based joint optimization of image compression and computer vision.
\newblock \emph{Neurocomputing}, 500: 13--25.

\bibitem[{Lin et~al.(2014)Lin, Maire, Belongie, Hays, Perona, Ramanan, Doll{\'a}r, and Zitnick}]{lin2014microsoft}
Lin, T.-Y.; Maire, M.; Belongie, S.; Hays, J.; Perona, P.; Ramanan, D.; Doll{\'a}r, P.; and Zitnick, C.~L. 2014.
\newblock Microsoft coco: Common objects in context.
\newblock In \emph{Proceedings of the European Conference on Computer Vision}, 740--755.

\bibitem[{Liu, Zhang, and Xiong(2019)}]{liu2019classification}
Liu, D.; Zhang, H.; and Xiong, Z. 2019.
\newblock On the classification-distortion-perception tradeoff.
\newblock In \emph{Proceedings of the International Conference on Neural Information Processing Systems}, 1206--1215.

\bibitem[{Liu, Sun, and Katto(2022)}]{liu2022improving}
Liu, J.; Sun, H.; and Katto, J. 2022.
\newblock Improving multiple machine vision tasks in the compressed domain.
\newblock In \emph{Proceedings of the IEEE International Conference on Pattern Recognition}, 331--337.

\bibitem[{Liu, Sun, and Katto(2023)}]{liu2023learned}
Liu, J.; Sun, H.; and Katto, J. 2023.
\newblock Learned image compression with mixed transformer-cnn architectures.
\newblock In \emph{Proceedings of the IEEE on Conference on Computer Vision and Pattern Recognition}, 14388--14397.

\bibitem[{Liu et~al.(2021)Liu, Liu, Li, Yan, and Li}]{liu2021semantics}
Liu, K.; Liu, D.; Li, L.; Yan, N.; and Li, H. 2021.
\newblock Semantics-to-signal scalable image compression with learned revertible representations.
\newblock \emph{International Journal of Computer Vision}, 129(9): 2605--2621.

\bibitem[{Liu et~al.(2023)Liu, Hu, Chen, and Xu}]{liu2023icmh}
Liu, L.; Hu, Z.; Chen, Z.; and Xu, D. 2023.
\newblock Icmh-net: Neural image compression towards both machine vision and human vision.
\newblock In \emph{Proceedings of the ACM International Conference on Multimedia}, 8047--8056.

\bibitem[{Minnen, Ball{\'e}, and Toderici(2018)}]{minnen2018joint}
Minnen, D.; Ball{\'e}, J.; and Toderici, G.~D. 2018.
\newblock Joint autoregressive and hierarchical priors for learned image compression.
\newblock \emph{Advances in Neural Information Processing Systems}, 31.

\bibitem[{Minnen and Singh(2020)}]{minnen2020channel}
Minnen, D.; and Singh, S. 2020.
\newblock Channel-wise autoregressive entropy models for learned image compression.
\newblock In \emph{Proceedings of the IEEE International Conference on Image Processing}, 3339--3343.

\bibitem[{Radford et~al.(2021)Radford, Kim, Hallacy, Ramesh, Goh, Agarwal, Sastry, Askell, Mishkin, Clark et~al.}]{radford2021learning}
Radford, A.; Kim, J.~W.; Hallacy, C.; Ramesh, A.; Goh, G.; Agarwal, S.; Sastry, G.; Askell, A.; Mishkin, P.; Clark, J.; et~al. 2021.
\newblock Learning transferable visual models from natural language supervision.
\newblock In \emph{Proceedings of the International Conference on Machine Learning}, 8748--8763.

\bibitem[{Redmon and Farhadi(2018)}]{redmon2018yolov3}
Redmon, J.; and Farhadi, A. 2018.
\newblock Yolov3: An incremental improvement.
\newblock \emph{arXiv preprint arXiv:1804.02767}.

\bibitem[{Shen et~al.(2024)Shen, Yin, Wang, He, Wang, and Yang}]{shen2024image}
Shen, X.; Yin, K.; Wang, X.; He, Y.; Wang, S.; and Yang, W. 2024.
\newblock Image Coding for Analytics via Adversarially Augmented Adaptation.
\newblock In \emph{Proceedings of the IEEE International Conference on Acoustics, Speech and Signal Processing}, 3605--3609.

\bibitem[{Shindo et~al.(2024)Shindo, Yamada, Watanabe, and Watanabe}]{shindo2024image}
Shindo, T.; Yamada, K.; Watanabe, T.; and Watanabe, H. 2024.
\newblock Image Coding for Machines with Edge Information Learning Using Segment Anything.
\newblock \emph{arXiv preprint arXiv:2403.04173}.

\bibitem[{Sullivan et~al.(2012)Sullivan, Ohm, Han, and Wiegand}]{sullivan2012overview}
Sullivan, G.~J.; Ohm, J.-R.; Han, W.-J.; and Wiegand, T. 2012.
\newblock Overview of the high efficiency video coding ({HEVC}) standard.
\newblock \emph{IEEE Transactions on Circuits and Systems for Video technology}, 22(12): 1649--1668.

\bibitem[{Torfason et~al.(2018)Torfason, Mentzer, Agustsson, Tschannen, Timofte, and Van~Gool}]{torfason2018towards}
Torfason, R.; Mentzer, F.; Agustsson, E.; Tschannen, M.; Timofte, R.; and Van~Gool, L. 2018.
\newblock Towards image understanding from deep compression without decoding.
\newblock \emph{arXiv preprint arXiv:1803.06131}.

\bibitem[{Wang et~al.(2021{\natexlab{a}})Wang, Wang, Yang, Zhang, Wang, Ma, and Gao}]{wang2021towards}
Wang, S.; Wang, S.; Yang, W.; Zhang, X.; Wang, S.; Ma, S.; and Gao, W. 2021{\natexlab{a}}.
\newblock Towards analysis-friendly face representation with scalable feature and texture compression.
\newblock \emph{IEEE Transactions on Multimedia}, 24: 3169--3181.

\bibitem[{Wang et~al.(2021{\natexlab{b}})Wang, Wang, Wang, and Ye}]{wang2021end}
Wang, S.; Wang, Z.; Wang, S.; and Ye, Y. 2021{\natexlab{b}}.
\newblock End-to-end compression towards machine vision: Network architecture design and optimization.
\newblock \emph{IEEE Open Journal of Circuits and Systems}, 2: 675--685.

\bibitem[{Wang et~al.(2022)Wang, Yu, De~Mello, Kautz, Anandkumar, Shen, and Alvarez}]{wang2022freesolo}
Wang, X.; Yu, Z.; De~Mello, S.; Kautz, J.; Anandkumar, A.; Shen, C.; and Alvarez, J.~M. 2022.
\newblock Freesolo: Learning to segment objects without annotations.
\newblock In \emph{Proceedings of the IEEE Conference on Computer Vision and Pattern Recognition}, 14176--14186.

\bibitem[{Xie, Cheng, and Chen(2021)}]{xie2021enhanced}
Xie, Y.; Cheng, K.; and Chen, Q. 2021.
\newblock Enhanced invertible encoding for learned image compression.
\newblock In \emph{Proceedings of the ACM International Conference on Multimedia}, 162--170.

\bibitem[{Yang et~al.(2021)Yang, Hu, Yang, Duan, and Liu}]{yang2021towards}
Yang, S.; Hu, Y.; Yang, W.; Duan, L.-Y.; and Liu, J. 2021.
\newblock Towards coding for human and machine vision: Scalable face image coding.
\newblock \emph{IEEE Transactions on Multimedia}, 23: 2957--2971.

\bibitem[{Yang et~al.(2024)Yang, Huang, Hu, Duan, and Liu}]{10440522}
Yang, W.; Huang, H.; Hu, Y.; Duan, L.; and Liu, J. 2024.
\newblock Video Coding for Machines: Compact Visual Representation Compression for Intelligent Collaborative Analytics.
\newblock \emph{IEEE Transactions on Pattern Analysis amp; Machine Intelligence}, 46(07): 5174--5191.

\bibitem[{Zhang et~al.(2018)Zhang, Isola, Efros, Shechtman, and Wang}]{zhang2018unreasonable}
Zhang, R.; Isola, P.; Efros, A.~A.; Shechtman, E.; and Wang, O. 2018.
\newblock The unreasonable effectiveness of deep features as a perceptual metric.
\newblock In \emph{Proceedings of the IEEE Conference on Computer Vision and Pattern Recognition}, 586--595.

\bibitem[{Zhong et~al.(2022)Zhong, Yang, Zhang, Li, Codella, Li, Zhou, Dai, Yuan, Li et~al.}]{zhong2022regionclip}
Zhong, Y.; Yang, J.; Zhang, P.; Li, C.; Codella, N.; Li, L.~H.; Zhou, L.; Dai, X.; Yuan, L.; Li, Y.; et~al. 2022.
\newblock Regionclip: Region-based language-image pretraining.
\newblock In \emph{Proceedings of the IEEE Conference on Computer Vision and Pattern Recognition}, 16793--16803.

\end{thebibliography}

\end{document}